# A*HAR: A NEW BENCHMARK TOWARDS SEMI-SUPERVISED LEARNING FOR CLASS IMBALANCED HUMAN ACTIVITY RECOGNITION

*Govind Narasimman*⋆    *Kangkang Lu*⋆    *Arun Raja*†    *Chuan Sheng Foo*⋆
Mohamed Sabry Aly◇    Jie Lin⋆    Vijay Chandrasekhar⋆

⋆ Institute for Infocomm Research (I²R), A∗STAR, Singapore
{govind_narasimman, lu_kangkang, foo_chuan_sheng, lin-j, vijay}@i2r.a-star.edu.sg
† The University of Edinburgh    A.R.-@sms.ed.ac.uk
◇ Nanyang Technological University, Singapore    msabry@ntu.edu.sg

## ABSTRACT

Despite the vast literature on Human Activity Recognition (HAR) with wearable inertial sensor data, it is perhaps surprising that there are few studies investigating semi-supervised learning for HAR, particularly in a challenging scenario with class imbalance problem. In this work, we present a new benchmark, called A*HAR, towards semi-supervised learning for class-imbalanced HAR. We evaluate state-of-the-art semi-supervised learning method on A*HAR, by combining Mean Teacher and Convolutional Neural Network. Interestingly, we find that Mean Teacher boosts the overall performance when training the classifier with fewer labeled samples and large amount of unlabeled samples, but the classifier falls short in handling unbalanced activities. These findings lead to an interesting open problem, i.e., development of semi-supervised HAR algorithms that are class-imbalance aware without any prior knowledge on the class distribution for unlabeled samples. The dataset and benchmark evaluation are released at https://github.com/I2RDL2/ASTAR-HAR for future research.

## 1. INTRODUCTION

Human Activity Recognition (HAR) based on inertial sensors have made considerable progress with advent of deep-learning algorithms and various databases [1], leading to development of health systems, elderly fall detection and sleep-quality monitoring, etc [2]. However, data collection and annotation have always been an issue in real applications. Both are time consuming and costly, HAR is no exception. Recent work has attempted to reduce the labelling effort required for HAR, by leveraging semi-supervised learning which is capable of training efficient classifiers using only few labeled samples and large amount of unlabeled samples [1].

Despite the promising results achieved on semi-supervised learning based HAR, difficulty in handling skewed activity distribution in dataset are yet to be completely addressed [3]. Daily activities in human life like drinking, combing-hair, door-opening, lying down and falling are sparse and shorter. Difficulty in manually labeling these rare events lead to much less number of samples for these sporadic classes, compared to other periodic activities with huge number of samples such as walking and eating. This will cause the well-known class-imbalance problem, which may easily drive a machine learning technique to predict activities with lesser data samples as a class with a large number of training instances [4]. Methods such as under-sampling majority class samples and over-sampling minority class samples can alleviate the class-imbalance problem [5], but prior knowledge on class distribution is presumed to be available which may not be the case when unlabeled samples present.

In this work, we develop a new benchmark A*HAR, towards exploring semi-supervised learning for class imbalanced HAR. We collected HAR data on a wide range of daily human activities ranging from short-term activities like FallStand and LyingDownGetUP to long-term activities like jogging and walking, using wrist wearable Mbientlab's Metatracker that contains different types of sensors including accelerometer, gyroscope, barometer and ambient light. As shown in Fig. 1 and Table 1, besides the high diversity of activities, A*HAR also exhibits high and varying degrees of class imbalance, with a standard derivation of 3.2 in terms of class distribution.

Numerous semi-supervised learning methods [6, 7, 8, 5] have been proposed on existing HAR datasets, but none of them explored the Mean Teacher method which has been widely recognized as a de-facto semi-supervised learning

---
Govind Narasimman, Arun Raja and Kangkang Lu contributed equally. This work is supported by the Agency for Science, Technology and Research (A∗STAR) under its AME Programmatic Funds (Project No.A1892b0026). Corresponding authors: Mohamed Sabry Aly, Jie Lin and Vijay Chandrasekhar.

framework recently. In this paper, we utilize A*HAR dataset to perform comprehensive HAR benchmarking of the Mean Teacher method under different configurations in varied challenging settings including training with or without unlabeled samples and training with balanced/unbalanced samples.

The contributions of the paper are:

- A new benchmark A*HAR enables the community to study HAR in more challenging real-world scenario that daily activities are highly imbalanced and labeled samples are limited, supplementing current state-of-the-art HAR datasets.

- An extensive semi-supervised learning based HAR evaluation is demonstrated on the A*HAR dataset under different configurations designed for evaluating the effects of training with labeled/unlabeled and balanced/unbalanced samples. Insights drawn from our performance analysis suggest interesting and open question for future research.

## 2. RELATED WORK

### 2.1. HAR Datasets

There are wide range of datasets publicly available for human activity recognition. Datasets like Darmstadt Daily Routines [9] and OPPURTUNITY [10] concentrate on daily chores, on the other hand datasets like UCI-HAR [11] and SP-HAD [12] emphasise on daily exercise and sports related activities. There are also HAR datasets [13, 14, 12] that exhibit class-imbalance, where some activities have very low amount of samples in dataset. UJAmi [13] has $\leq 100$ samples for washing dishes and entering room. SP-HAD [12] has only 2% samples for jumping.

As summarized in Table 1 and Fig. 1, the A*HAR dataset also has heavy class-imbalance in distribution of activities data. A*HAR contains 24 activities with a wide range and category of activities which includes Activities of Daily Living, high intensity exercises and light exercises. The dataset exhibits high imbalance between maximum and minimum cardinality activity (i.e., Max-Min ratio 28). The distribution of activities in dataset have a standard deviation ($\sigma$) of 3.2. PAMAP2 which contains both Activities of Daily life and exercises have a smaller Max-Min Ratio of 6 and Std. of 2.62, whereas SP-HAD and UJAmi are constrained to one category of activities and contains less number of classes.

### 2.2. Semi-supervised Learning for HAR

Semi-supervised learning usually can boost the application accuracy significantly by leveraging limited labeled samples and large amount of unlabeled samples. Consistency-based method has been one of the most widely used semi-supervised learning methods. Rasmus and Valpola [15] proposed ladder network which can support unsupervised learning by encoder and decoder mapping. Laine and Aila [16] proposed Temporal Ensembling to enforce the consistency between predictions across different epochs. In [17], the moving average of weights is further used to build a teacher model with more confident predictions. The consistency loss between student and teacher model boost the performance to a large margin comparing with previous methods.

**Table 1**. Comparison of HAR datasets that exhibit class-imbalance problem. Std. - Standard deviation of the distribution of activities. Max-Min R. - Ratio between maximum and minimum cardinality activity. #Act. - Number of activities. EXE - Exercise, ADL - Activities of Daily Living.

| Dataset | Std. | Max-Min R. | #Act. | Type |
|---|---|---|---|---|
| PAMAP2 [14] | 2.62 | 6 | 18 | EXE,ADL |
| UJAmi [13] | 3.03 | 69 | 24 | ADL |
| SP-HAD [12] | 7.41 | 14 | 9 | EXE |
| A*HAR (ours) | 3.2 | 28 | 24 | EXE,ADL |

Zeng and Yu [6] firstly applied ladder network to human activity recognition for semi-supervised learning. Lv and Chen [7] trained two classifiers with two views and used co-training to make classifiers cooperate on unlabeled data. Zhu and Chen [8] proposed to learn high-level features using long short-term memory and utilised the unlabeled data by ensembling predictions in the past epochs. Chen and Yao [5] used multi-model to assign pseudo labels to unlabeled data. Besides, they also tried to solve the class imbalance problem of labeled data by over-sampling and under-sampling samples in the feature space. None of these researches have evaluated the Mean Teacher method which is used as the baseline for most of the recent semi-supervised learning algorithms.

**Remark**. To the best of our knowledge, the new benchmark A*HAR developed in this work is the first to evaluate the state-of-the-art semi-supervised method, Mean Teacher, in the context of HAR with class-imbalance problem. More importantly, there is an interesting research question emerged from the evaluation results for future study - development of Mean Teacher based HAR algorithms that are class-imbalance aware. In practice, we can easily see the class distribution of the labeled data. However, for unlabeled data, class distribution is unknown. Thus, the main challenge is how to adapt the Mean Teacher method to imbalanced HAR dataset without any prior knowledge on the class distribution for unlabeled samples.

## 3. THE A*HAR DATASET

The user-friendly and cost efficient wearable sensor, Mbientlab-Metatracker with Accelerometer (A), Gyroscope (G), Barometer(P) and ambient light sensor(L) is used to collect inertial data, with the sensor tied to the wrist of subject. For A*HAR, the total number of activities is 24, including high intensity exercises like Dumbbell, Weight-Lifting, Squats, Push-Up, Crunches, Sit-Ups and Sprinting, low intensity exercises like Jogging, Cycling-Indoor, Climbing-Up, Climbing-Down, Biking, Walking and Lunges. Sensor data was also collected for daily activities like Combing Hair, Lie Down-Get Up, Drinking, Fall-Stand, Eating, Brushing and Typing. These activities can be also categorized into short-term activities like FallStand and LyingDownGetUP and long-term activities like jogging and walking.

**Table 2**. Statistics of the raw data collected by 9 subjects.

| Sensor | Frequency (Hz) | # Readings |
|---|---|---|
| Accelerometer (A) | 25 | 795,206 |
| Gyroscope (G) | 25 | 795,206 |
| Barometer (P) | 32.5 | 1,033,768 |
| Ambient Light (L) | 10 | 318,082 |

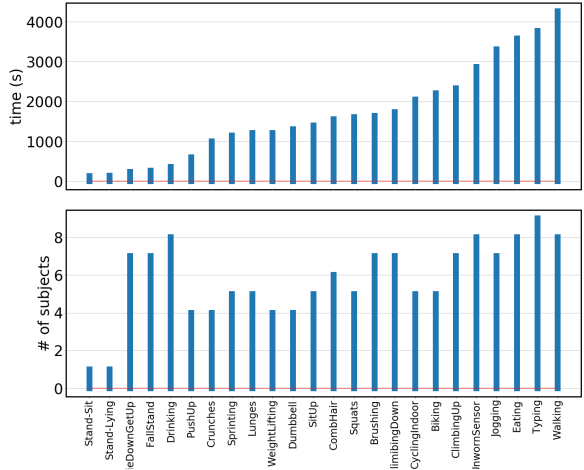

**Fig. 1**. (a) Amount of data collected per activity (in seconds) and (b) Number of subjects per activity.

Activities are carried out by 9 subject aged between 21-45 with a height range of 160-178 cm. All subjects voluntarily collect a subset of prescribed activities in diverse environments according to their convenience. Table 2 summarizes data collection frequency and number of raw data readings for different sensors. The ground-truth annotations for activities are generated using the MetaBase app installed on mobile phone. For collecting data for each activity, one has to create a new session on the app, sensor data in CSV files format are created after the session is ended.

A*HAR exhibits highly imbalanced class distribution. Fig. 1 shows the distribution of amount of data collected per activity and number of subjects who conducted the activity. Sensor data during exercises like jogging and cycling were collected for 300s of approximate duration. Sensor data from daily activities that consume less time like Brushing Drinking and Combing- Hair etc amount to less than 60s. High intensity exercises such as PushUps, Crunches and Sprinting also have small number of samples.

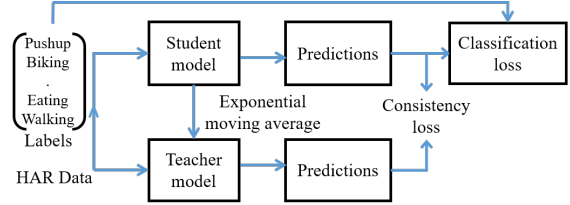

**Fig. 2**. Architecture of Mean Teacher framework for Semi-supervised Learning.

For data pre-processing, the raw data collected from the sensors are noise filtered and concatenated. We snap 2.5s fixed-width segments with 50% overlap ratio from the raw signals. Following [11], the data from Accelerometer was separated into body component and acceleration component. A data sample is formed by concatenating Accelerometer and Gyroscope values as a 2D array with size 64 x 9, in which 64 stands for the number of readings per segment. For each subject we split the data samples into training and testing, where the first 70% of the readings is added into training set and the remaining for testing. Finally, the total number of samples is 24,770, 7,520 out of them are fixed as testing set.

## 4. THE SEMI-SUPERVISED BASELINE

A framework figure of the semi-supervised based method, Mean Teacher, is shown in Figure 2. The Mean Teacher method [17] attempts to have consistent predictions despite of the small perturbations on both data and model parameters. In this method, we have a student and teacher network sharing the same convolutional neural network architecture. During training, both labeled and unlabeled samples are randomly selected as a mini-batch. These batches of data are then feed into both the student and teacher models. However different perturbations are applied to student and teacher models including Gaussian noise, flips, and dropout. For the labeled samples, the classification loss is computed as the cross-entropy between the predictions of student model and the ground truth. For the unlabeled data, the mean-square distance between the prediction from the student and teacher models are used as the consistency loss. In the optimization process, only the weights of student model is updated while the weights of the teacher model are the Exponential Moving Average (EMA) of the weights from student model.

## 5. EXPERIMENTS AND ANALYSIS

### 5.1. Experimental Setup

We apply the same architecture as used in Mean Teacher paper [17], a 13-layer convolution neural network. We slightly modified the network to fit the dimension of our data (i.e., 64 x 9). The model is trained using the ADAM optimizer with a batch size of 100 for 37,500 steps, The maximum learning rate is set to 0.003. All experiments are done five times with the same set of random seeds, and the classification accuracy is reported as the average of multiple runs along with standard deviations.

**Table 3**. Classification Accuracy (%) of Mean Teacher method without or with unlabeled samples during training on A*HAR. "All" stands for the supervised baseline using all labels during training.

| Labels | w/o Unlabeled | w Unlabeled |
|--------|---------------|-------------|
| 288    | 39.13 ± 2.14  | 41.79 ± 2.76 |
| 600    | 52.44 ± 1.68  | 59.93 ± 1.60 |
| 1200   | 67.96 ± 2.45  | 71.55 ± 1.31 |
| All    | 97.78 ± 0.07  | -           |

**Table 4**. Classification Accuracy (%) of Mean Teacher method without or with class-imbalance during training on a subset of A*HAR.

| Labels | Balanced   | Unbalanced |
|--------|------------|------------|
| 288    | 44.6 ± 1.1 | 42.8 ± 2.5 |
| 600    | 59.6 ± 1.2 | 58.6 ± 0.9 |
| 1200   | 73.0 ± 0.91| 69.1 ± 2.51|

### 5.2. Mean-Teacher with or without Unlabeled data

In this study, we evaluate the Mean Teacher method in 2 dataset settings: training with labeled samples only and training by adding unlabeled samples to the labeled subset. The labeled subset is randomly extracted from the full training set with 17,250 samples and balanced across the 24 activities. The remaining samples are used as unlabeled set in which the class distribution is imbalanced apparently.

Evaluation results on the fixed testing set are reported in Table 3. One can see that Mean Teacher consistently improves the classification accuracy by combining large amount of unlabeled samples with few labeled samples during training, regardless of the ratio of training labels used. Moreover, the performance gap between semi-supervised Mean Teacher and supervised baseline can be bridged by adding more labels into the training loop.

### 5.3. Mean-Teacher with or without Class-Imbalance

Though adding unlabeled samples improves the performance of Mean Teacher method, it leads to another important question that the effect of class distribution for both labeled and unlabeled samples is unclear. To answer this question, we design an experiment where the Mean Teacher is trained with 2 dataset settings: training with a Balanced subset in which the class distribution is balanced for both labeled and unlabeled samples, and training with a Unbalanced subset with the class distribution of A*HAR well respected for both labeled and unlabeled samples. One may note that 5 activities with the lowest cardinalities are removed as they are not sufficient for the construction of the subsets. Finally, the total number of training samples for both Balanced and Unbalanced subsets is 4,200.

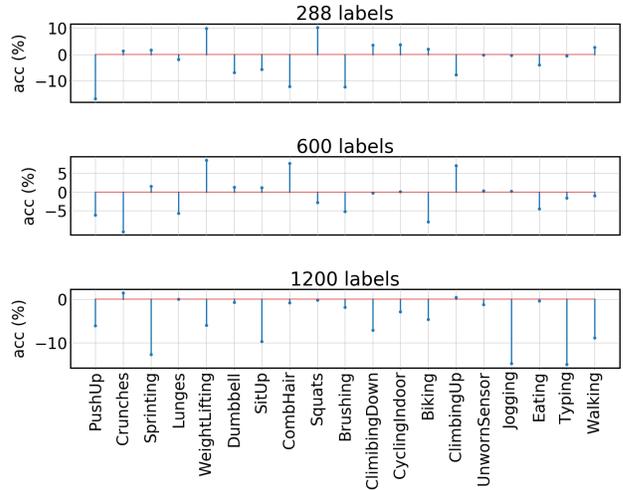

**Fig. 3**. Activity-wise classification accuracy differences by subtracting class-balanced Mean Teacher performance from class-unbalanced Mean Teacher performance.

Table 4 shows the evaluation results on the fixed testing set. In addition, Fig. 3 illustrates the break-down of activity-wise classification accuracy differences by subtracting class-balanced Mean Teacher performance from class-unbalanced Mean Teacher performance. We observe that Mean Teacher with Unbalanced subset performs consistently worse than Mean Teacher with Balanced subset for most of the activities, for varying ratios of training labels used. The performance drop is probably due to the reason that Mean Teacher is naturally not aware of the unbalanced class distribution, in particularly for the large amount of unlabeled samples.

### 6. CONCLUSION

In this paper, we present a new benchmark, A*HAR, for semi-supervised learning based HAR with heavy class imbalance problem. We found that the Mean Teacher method boosts performance when adding unlabeled samples to the relatively fewer labelled samples during training. However, it degrades in performance when handling class-unbalanced HAR where the number of samples are varying across activities. Thus, we expect that A*HAR, as a step forward to HAR in real-world scenario, will drive rapid progress on development of new semi-supervised Mean Teacher algorithms that are robust to unbalanced class distributions.